%% file: 00root.tex
\documentclass[letterpaper, 10 pt, conference]{ieeeconf}  

\IEEEoverridecommandlockouts                              

\usepackage{graphics} 
\usepackage{epsfig} 
\usepackage{times} 
\usepackage{amsmath} 
\usepackage{amssymb}  
\usepackage{multirow}  
\usepackage{gensymb}
\usepackage{subcaption}
\usepackage{graphicx}
\usepackage{tikz}

\usepackage[colorlinks=true]{hyperref}

\title{\LARGE \bf
CalibNet: Geometrically Supervised Extrinsic Calibration using 3D Spatial Transformer Networks
}

\author{Ganesh Iyer, R. Karnik Ram, J. Krishna Murthy, and K. Madhava Krishna
\thanks{Ganesh Iyer, R. Karnik Ram, and K. Madhava Krishna are with the Robotics Research Center at the International Institute of Information Technology, Hyderabad, India.
        J. Krishna Murthy is with Mila, Universit\'e de Montr\'eal, Quebec, Canada.
        Authors' email: \tt{\small{giyer2309@gmail.com, karnikram@gmail.com, mkrishna@iiit.ac.in}}}%
}

\newcommand\copyrighttext{%
  \footnotesize \textcopyright 2018 IEEE. Personal use of this material is permitted.
  Permission from IEEE must be obtained for all other uses, in any current or future
  media, including reprinting/republishing this material for advertising or promotional
  purposes, creating new collective works, for resale or redistribution to servers or
  lists, or reuse of any copyrighted component of this work in other works.
  DOI: \href{https://doi.org/10.1109/IROS.2018.8593693}{10.1109/IROS.2018.8593693}}
\newcommand\copyrightnotice{%
\begin{tikzpicture}[remember picture,overlay]
\node[anchor=south,yshift=10pt] at (current page.south) {\fbox{\parbox{\dimexpr\textwidth-\fboxsep-\fboxrule\relax}{\copyrighttext}}};
\end{tikzpicture}%
}

\begin{document}

\makeatletter
\let\@oldmaketitle\@maketitle
\renewcommand{\@maketitle}{\@oldmaketitle
\centering
\includegraphics[width=0.85\textwidth]{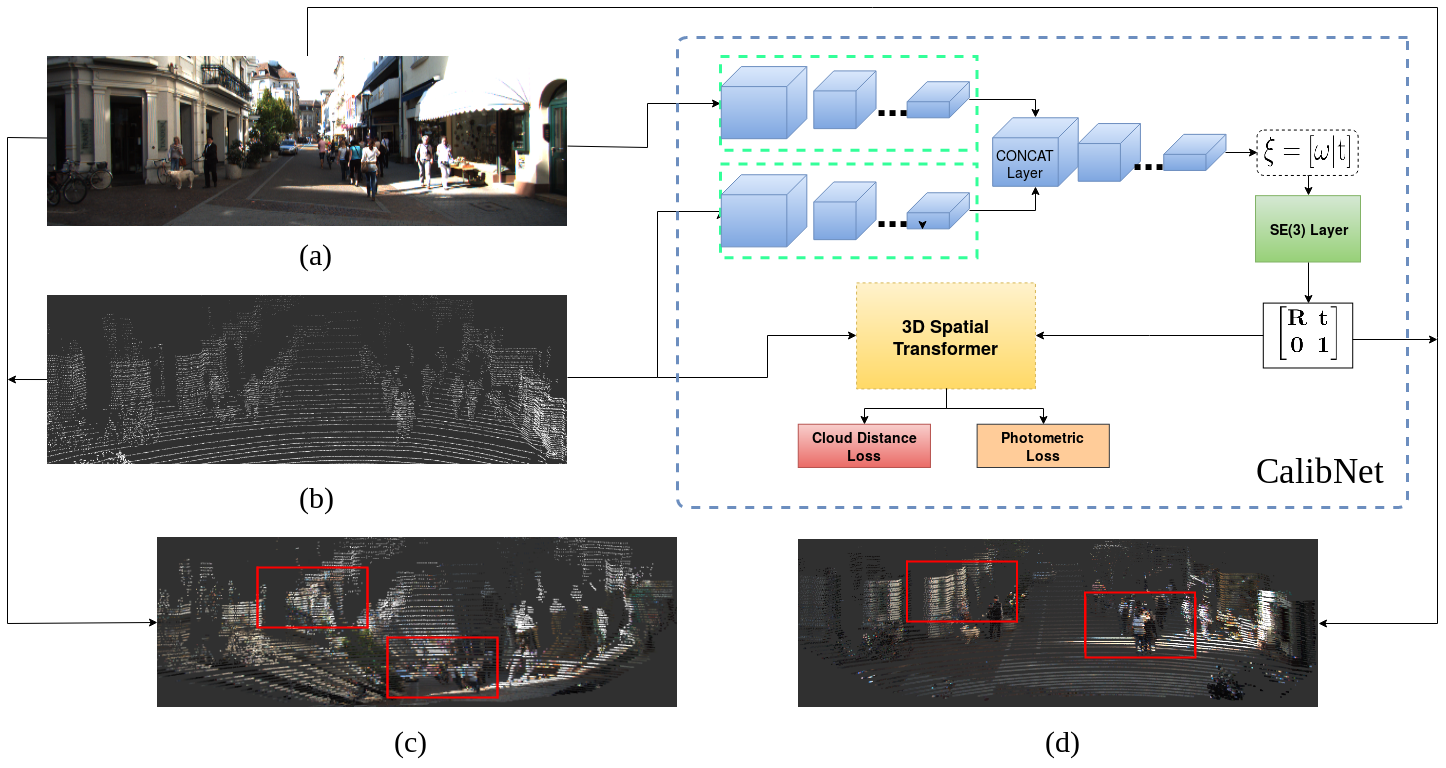}
\vspace{0.2cm}
\captionof{figure}{CalibNet estimates the extrinsic calibration parameters between a 3D LiDAR and a 2D camera. It takes as input an RGB image (a) from a calibrated camera, a raw LiDAR point cloud (b), and outputs a $6$-DoF rigid-body transformation $T$ that \emph{best} aligns the two inputs. (c) shows the colorized point cloud output for a mis-calibrated setup, and (d) shows the output after calibration using our network. As shown, using the mis-calibrated point cloud to recover a colorized 3D map of the world results in an incoherent reconstruction. Notice how the 3D structures highlighted in (c) using red rectangles fail to project to their 2D counterparts. However, using the extrinsic calibration parameters predicted by CalibNet produces more consistent and accurate reconstructions (d), even for large initial mis-calibrations.}}
\label{fig:teaser}
\makeatother
\hypersetup{urlcolor=black}
\maketitle
\copyrightnotice
\thispagestyle{empty}
\pagestyle{empty}

\begin{abstract}

3D LiDARs and 2D cameras are increasingly being used alongside each other in sensor rigs for perception tasks. Before these sensors can be used to gather meaningful data, however, their extrinsics (and intrinsics) need to be accurately calibrated, as the performance of the sensor rig is extremely sensitive to these calibration parameters. A vast majority of existing calibration techniques require significant amounts of data and/or calibration targets and human effort, severely impacting their applicability in large-scale production systems. We address this gap with CalibNet: a geometrically supervised deep network capable of automatically estimating the 6-DoF rigid body transformation between a 3D LiDAR and a 2D camera in real-time. CalibNet alleviates the need for calibration targets, thereby resulting in significant savings in calibration efforts. During training, the network only takes as input a LiDAR point cloud, the corresponding monocular image, and the camera calibration matrix K. At train time, we do not impose direct supervision (i.e., we do not directly regress to the calibration parameters, for example). Instead, we train the network to predict calibration parameters that maximize the geometric and photometric consistency of the input images and point clouds. CalibNet learns to iteratively solve the underlying geometric problem and accurately predicts extrinsic calibration parameters for a wide range of mis-calibrations, without requiring retraining or domain adaptation. The project page is hosted at \url{https://epiception.github.io/CalibNet}


\end{abstract}

\hypersetup{
    colorlinks=true,   
    urlcolor=black,
    linkcolor=black
}


\input{01intro}

\input{02relatedwork}

\input{03approach}

\input{04results}

\input{05conclusion}







\small{
\bibliographystyle{IEEEtran}
\bibliography{references}
}

\end{document}

%% file: 01intro.tex
\section{INTRODUCTION}

Perception of visual cues and objects in the environment is an important aspect of autonomous robot navigation. A successful perception system relies on various on-board sensors. A growing number of sensors of various modalities are being used in robots. An autonomous car, for instance, uses a 3D LiDAR
in combination with 2D cameras as the dense color information of the latter complements the sparse distance information of the former.
With their increasing use, calibration techniques to estimate accurate extrinsic parameters of the sensors are becoming increasingly important. 
In the same example of an autonomous car,
without accurate extrinsic parameters, the laser distance measurements cannot be accurately projected onto the camera images, and thus color pixels
in the images cannot be accurately associated with distance information. Over the past several years, a number of calibration techniques have been proposed, specifically for the LiDAR-camera calibration problem \cite{unnikrishnan2005fast}, \cite{geiger2012automatic}, \cite{levinson2013automatic}, \cite{pandey2012automatic}, \cite{taylor2015motion}
Yet, the vast majority of these techniques depend on specific calibration targets such as checkerboards, and require significant
amounts of manual effort \cite{unnikrishnan2005fast}, \cite{geiger2012automatic}. In addition, they are unable to correct for any deviation due to environmental changes or vibrations during live operation, often rendering the robots inoperable.
Thus there is an imperative need for automatic and online calibration techniques which can significantly extend the flexibility and adaptability of these robots. 
There have been some previously published techniques in this area \cite{levinson2013automatic}, \cite{pandey2012automatic}, \cite{taylor2015motion} but most of these techniques still depend on accurate initialization for the calibration parameters \cite{levinson2013automatic}, \cite{pandey2012automatic}, or require signficant amounts of ego-motion data \cite{taylor2015motion}.

Our work tries to tackle this problem of LiDAR-camera calibration i.e. 
estimating the $6$-DoF rigid body transformation between a 3D LiDAR and a 2D camera, without any assumptions about the existence of any specific features or landmarks
in the scenes, without any initial estimate for the extrinsic parameters, and in real-time.
We leverage the recent success of deep neural networks in classical computer vision tasks such as
visual recognition \cite{krizhevsky2012imagenet}, in our solution. 
Our network only takes as input a LiDAR point cloud, the corresponding monocular image, and the camera calibration matrix $K$, and is able to accurately estimate the extrinsic parameters for 
a wide range of possible mis-calibrations about and along any of the axes respectively. Further, our training method employs geometric supervision by directly reducing the dense photometric error and dense point cloud distance error measures, to regress the correct extrinsic calibration parameters, thus eliminating the need of an existing calibrated sensor setup to collect training data.
To the best of our knowledge, we believe this is the first deep neural network to estimate extrinsic calibration parameters in this manner. 

We showcase the following contributions with \textit{CalibNet}:
\begin{enumerate}
\item CalibNet is the first geometrically supervised deep learning approach to tackle the problem of multi-sensor self-calibration in LiDAR-camera rigs.
\item We use a novel architecture based on 3D Spatial Transformers \cite{handa2016gvnn} that learns to solve the underlying physical problem, by using geometric and photometric consistency to guide the learning process. This makes CalibNet agnostic to non-invariant parameters such as dependence on camera intrinsics, and the approach effortlessly generalizes to data collected from multiple sensor rigs, without requiring retraining or fine-tuning.
\item In the larger scheme of things, the paper presents a fundamentally different approach for model-to-image registration to leverage the best of both worlds (viz. deep representation learning and geometric computer vision) in building an accurate and viable system for the task.
\end{enumerate}

The paper is organized as follows: We review related LiDAR-camera calibration techniques in the next section. We detail our
technique and the network architecture in Sec. III. In Sec. IV we experimentally evaluate our technique, 
and conclude the paper in Sec. V.

%% file: 02relatedwork.tex
\section{RELATED WORK}

\begin{figure*}
\begin{center}
	\includegraphics[width=1.0\linewidth]{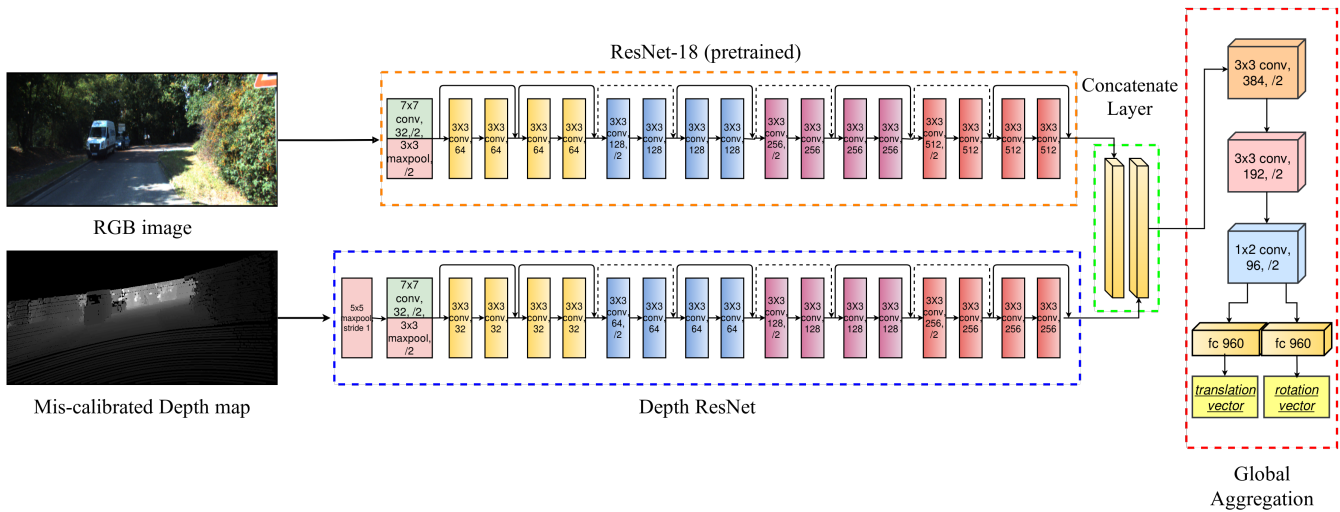}
\end{center}
	\caption{Network architecture}
\label{fig:overall_net}
\end{figure*}

The LiDAR-camera extrinsic calibration problem has been well-studied for several years. Existing approaches for cross-sensor calibration broadly fall into two categories, viz. target-based and target-less techniques. The taxonomy can be extended further based on whether they work automatically
or require manually labeled correspondences. Usually, there is a trade-off between the effort required in experimental setup and the amount of data needed for effective calibration. Expensive calibration setups have the advantage that they work with very little data. Inexpensive calibration  setups compensate for the lack of sophistication by gathering a larger volume of data.

Geiger et al. \cite{geiger2012automatic} proposed an automatic system for accurate camera-camera and LiDAR-camera calibration using just a single image per sensor. They require a specific calibration setup with multiple checkerboard targets. 
Simpler techniques that can solve for the LiDAR-camera extrinsic parameters, using simple easy-to-make targets and lesser number of correspondences have been recently proposed in \cite{beltran2017lidar},\cite{pusztai2017accurate}.
Although accurate, these techniques are slow, labor intensive, and require careful tuning of several hyperparameters.

Levinson and Thrun proposed one of the first target-less techniques in \cite{levinson2013automatic}. Their underlying assumption, is that depth discontinuities in laser data should project onto edges in images for an accurate extrinsic calibration. The two feature classes are
combined and a grid search is used to estimate the optimal extrinsic parameters. Pandey et al. proposed a very similar method in \cite{pandey2012automatic} 
where they try to maximize the mutual information between the LiDAR's intensity of return and the intensity of the corresponding points in the camera's image. Although the assumptions hold even in the presence of measurement noise, the formulated cost functions are only locally convex and hence rely on a good initialization for the optimization to converge.

Another class of target-less techniques exists where independent motion estimates from the sensors are used and matched to obtain the extrinsic
calibration parameters, as shown in \cite{taylor2015motion}. They do not rely on any initialization for the extrinsics, or require any overlapping
fields of view for the sensors. However, they still require large amounts of data and good motion estimates for accurate results which limits their applicability to offline scenarios.

Recently, deep neural networks have shown tremendous success in classical computer vision tasks such as visual recognition \cite{krizhevsky2012imagenet}, localization \cite{kendall2015posenet},
and correspondence estimation \cite{choy_nips16}. Deep networks have also shown their effectiveness in dealing with unordered point clouds for tasks such as 3D Object Detection and Segmentation \cite{qi2017pointnet}, \cite{qi2017pointnet++}.
Yet, surprisingly, only a few deep learning based approaches have been applied to the calibration problem. The first deep convolutional neural network for LiDAR-camera calibration was
proposed in \cite{schneider2017regnet} by Schneider et al. Using a Network-in-Network based pre-trained supervised network, they aim to regress the transformation parameters that accurately aligns the LiDAR point cloud to the image, by training the network with large amounts of annotated ``decalibration'' data.
While feasible and real-time, the training process is agnostic of the underlying geometry of the problem. Schneider et al. \cite{schneider2017regnet} regress to the calibration parameters, conditioned on the input image. Since it doesn't take geometry into account, it has to be retrained each time sensor intrinsics change. In contrast, our method leverages the recent success of self-supervised networks \cite{zhou2017unsupervised}, \cite{li2017undeepvo} 
and attempts to solve the problem by attempting to reduce the dense photometric error and dense point cloud distance error between the misaligned and target depth maps.
While we use transformed depth maps as targets, such a map could be found by any stereo reconstruction method and be used for training. 
Further, since our model only requires camera intrinsics for applying the spatial transformations during training, any intrinsically calibrated camera system can be used for extrinsic calibration using our architecture.

%% file: 03approach.tex
\section{OUR APPROACH}

In this section we present the theory behind our approach, the network architecture, training methodology, and loss functions.

    

\subsection{Network Architecture} 

    

\textbf{\textit{Input Preprocessing:}} The network takes as input an RGB image, the corresponding mis-calibrated LiDAR point cloud, and the camera calibration matrix $K$.

\begin{figure*}
\begin{center}
    \includegraphics[width=0.95\linewidth]{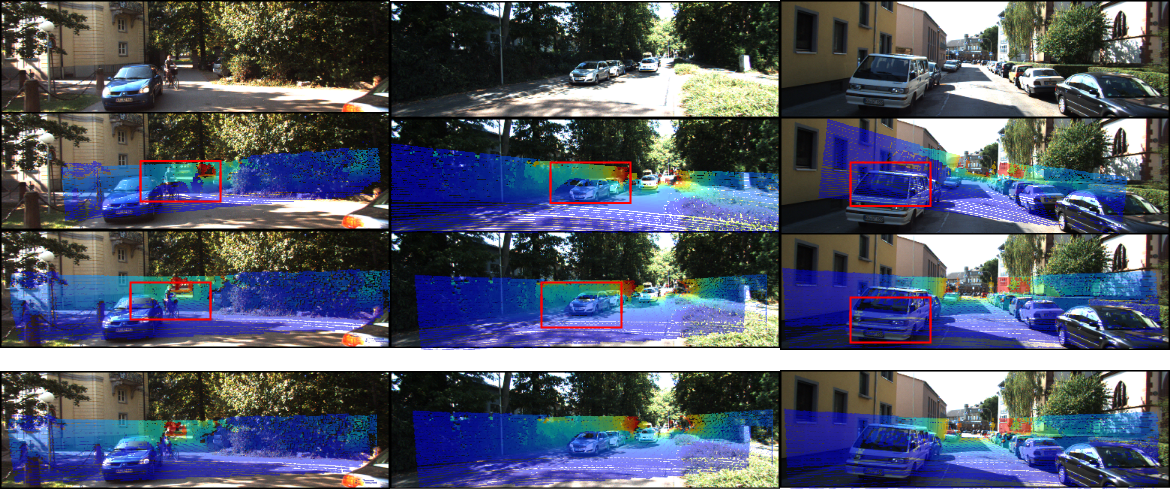}
\end{center}
	\caption{Examples of calibration results for different scenes from the KITTI dataset. The first row shows the input RGB images, and the second row shows the corresponding mis-calibrated LiDAR point cloud projected onto the images. The third row shows LiDAR point clouds projected using the network's predicted transforms, and the last row shows the corresponding ground truth results. The red rectangles in the second row indicate the mis-alignment, and in the third row they denote the proper alignment after calibration.}
\label{fig:long}
\end{figure*}

The point cloud is first converted into a sparse depth map as a pre-processing step. This is done by projecting the LiDAR point cloud onto the image plane. Since the initial mis-calibration is inaccurate, projecting the mis-calibrated points to the image plane results in a sparse depth map that is (grossly) inconsistent with the image (see Fig. 1(c)). We normalize both the RGB input image and the sparse depth map to the range of $\pm 1$. The sparse depth maps are then max-pooled to create semi-dense depth maps using a 5 x 5 max-pooling window. The resulting semi-dense depth maps look similar to the inputs shown in Fig.\ref{fig:long}(b).\\


\textbf{\textit{Architectural Details:}} The network primarily consists of $2$ asymmetric branches, each performing a series of convolutions (see Fig. \ref{fig:overall_net}. For the \emph{RGB branch} we use the convolutional layers of a pre-trained ResNet-18 network \cite{he2016deep}. For the \emph{depth branch}, we use a similar architecture as for the RGB stream, but with half the number of filters at each stage. Like in \cite{schneider2017regnet}, this architecture has several advantages for feature extraction.
The use of pre-trained weights for the RGB input prevents learning the relevant features from scratch. However, since the parameters of the depth stream are learned from scratch, the filters for the depth stream are reduced at each stage. 
The outputs of the two branches are then concatenated along the channel dimension and passed through a series of additional fully convolutional layers, for global feature aggregation. BatchNorm \cite{batchnorm} is used throughout the network, after every convolutional block. We decouple the output streams for rotations and translations to capture differences in modalities that might exist between rotations and translations. The output of the network is a 1 x 6 vector $\xi = (v, \omega) \in se(3)$ where $v$ is the translational velocity vector, and $\omega$ is the rotational velocity vector.

\subsubsection*{\textbf{SO(3) layer}} While translations are directly predicted at the output of the network, we need to convert the output rotation vector in $so(3)$ to its corresponding rotation matrix. An element $\omega \in so(3)$ can be converted to $SO(3)$ by using the Exponential Map. The exponential map is simply the matrix exponential over a linear combination of the group generators. Given $\omega = (\omega_1, \omega_2, \omega_3)^T$, the exponential map is defined as follows.
\begin{equation*}
exp: so(3) \rightarrow SO(3); \; \hat{\omega} \mapsto e^{\hat{\omega}}
\end{equation*}
Here, $\hat{\omega}$ is the skew-symmetric matrix from of $\omega$ (also referred to as the \emph{hat} operator), and $e^{\hat{\omega}}$ computed using Taylor series expansion for the matrix exponential function. A closed form solution to the above expression yields the well-known Rodrigues formula.

\begin{equation*}
R = e^{\hat{\omega}} = I + \frac{\hat{\omega}}{\Vert{\omega}\Vert}\sin{\Vert{\omega}\Vert} + \frac{\hat{\omega}^2}{\Vert{\omega}\Vert^2}(1 - \cos(\Vert\omega\Vert))
\end{equation*}

This gives us the rotation in \textbf{$R$} in $SO(3)$. Combining with translation predicted by the network gives us a  3D rigid body transformation $T \in SE(3)$ defined as
$$
   T = \left( \begin{array}{cc} R & t \\ 0 & 1 \end{array}\right) \text{where } R \in SO(3) \thinspace \text{and } t \in \mathbb{R}^3
$$

\textbf{\textit{3D Spatial Transformer Layer:}} Once we convert the calibration parameters predicted by the network to a rigid-body transform in $T \in SE(3)$, we use a 3D Spatial Transformer Layer that transforms the input depth map by the predicted transformation $T$. We extend the original 3D Spatial Transformer layer \cite{handa2016gvnn}  in this work to handle sparse or semi-dense input depth maps.



Knowing the camera intrinsics $(f_x, f_y, c_x, c_y)$ allows back-projection of the max-pooled depth image to a sparse (or in some cases, semi-dense) point cloud using the mapping $\mathbb{\pi}^{-1}:\mathbb{R}^2\rightarrow\mathbb{R}^3$.

\begin{equation*}
\mathbb{\pi}^{-1}(x,y,Z) = \left( \left(\dfrac{x - c_x}{f_x}\right), \left(\dfrac{y - c_y}{f_y}\right), Z \right)
\end{equation*}

We now transform the obtained point cloud by the extrinsic calibration $T$ predicted by the network, and then project the transformed point cloud back to the image plane using the camera intrinsics.

\begin{equation}
\left(\begin{matrix}x \\ y \end{matrix} \right) = \pi \left( R \left( \begin{matrix} X \\ Y \\ Z \end{matrix}  \right) + t \right)
\end{equation}
Here, $R$ and $t$ are the rotation and translation components of $t$, and $\pi$ is the perspective projection operator (note that the camera intrinsics are subsumed into $\pi$, for sake of brevity).

This operation is carried out in a differentiable manner using the 3D Grid Generator. To obtain an image of similar dimensions to that of the input image, we scale $(f_x, f_y, c_x, c_y)$ accordingly.

\textbf{\textit{Loss Functions:}} The use of dense methods for registration is even more requisite in the case of extrinsic calibration. We use two types of loss terms during training:\\

\textit{1. Photometric Loss:} After transforming the depth map by the predicted $T$, we check for the dense pixel-wise error (since each pixel is encoded with the depth intensity) between the predicted depth map and the correct depth map. The error term is defined as,

\begin{equation}
    \mathcal{L}_{photo} = {1 \over 2} \sum_{1}^{N}\left(D_{gt} - KT\pi^{-1}[D_{miscalib}] \right) ^2
\end{equation} 
where $D_{gt}$ is the target depth map and $D_{miscalib}$ is the initial mis-calibrated depth map. While we use max-pooled depth maps during training, note that this could be further generalized by using a stereo pair to estimate depth maps, and use the same for training as well.\\

\textit{2. Point Cloud Distance Loss:}
The 3D Spatial Transformer Layer allows the transformation of a point cloud after backprojection. At this stage, we utilize the unregistered transformed and target point clouds and try to minimize the 3D-3D point distances between them in metric scale. Note that we don't know correspondences since we are working with unordered point sets, making it a more difficult task. Therefore, we considered various distance metrics that would be an accurate measure of the error between unregistered sets in the world-coordinate frame. Following the recent success of \cite{fan2017point} for point cloud generation, we considered the following distance measures.\\

\subsubsection*{Chamfer Distance} The Chamfer Distance between two point clouds $S_1,S_2 \subseteq \mathbb{R}^3$, is defined as the sum of squared distances of the nearest points between the two clouds. 

\begin{equation}
    d_{CD}(S_1,S_2) = \sum_{x \in S_1} \min_{y \in S_2} \left \| x - y \right \|_{2}^{2} + \sum_{y \in S_2} \min_{x \in S_1} \left \| x - y \right \|_{2}^{2}
\end{equation} 

\subsubsection*{Earth Mover's Distance} The Earth Mover's Distance is originally a measure of dissimilarity between two multi-dimensional distributions. Since the distance between the individual points of the clouds can be calculated, we try to solve for a metric that signifies the overall distance measure between the clouds. Specifically, given two point clouds $S_1,S_2 \subseteq \mathbb{R}^3$, the optimization essentially tries to solve the assignment problem for each point. In particular, if $\phi$ is a mapping between the two point sets, then we minimize the distance as follows, 

\begin{equation}
    d_{EMD}(S_1,S_2) = \min_{\phi:S_1 \rightarrow S_2} \sum_{x \in S1} \left \| x - \phi(x) \right \|_{2}
\end{equation} 

where $\phi: S_1 \rightarrow S_2$ is a bijection. \\

\subsubsection*{Centroid ICP Distance} Similar to the loss term used for Iterative Closest Point based alignment, we try to directly minimize the distance between the target point clouds and a point cloud transformed by the predicted transformation. While various methods are used to establish an initial correspondence, such as the use of Kd-trees or minimum distance between points, we directly minimize the total distance between point cloud cluster centres. Each centre is computed as a centroid of intermittent points.   

\begin{equation}
    d_{icp}(S_1,S_2) = \dfrac{1}{2}\sum_{1}^{N} \sum_{1}^{i}\left \| X_{exp}^i - ({R}X_{miscalib}^i + {t}) \right \|^2
\end{equation} 
where $X_{exp}$ is a possible cluster center of the expected point cloud, and $X_{miscalib}$ is a cluster center from the mis-calibrated point cloud projected to the world coordinate frame.\\

Our final loss function consists of a weighted sum of the photometric loss and point cloud distance loss, 

\begin{equation}
    \mathcal{L}_{final} = \alpha_{ph}(\mathcal{L}_{photo}) + \beta_{dist}(d(S_1,S_2))
\end{equation}

\subsection{Layer Implementation Details}

A major bottleneck when dealing with semi-dense depth maps is that we cannot utilize operations that apply updates to all tensor locations. Since the input maps would contain multiple pixel locations with zero depth as intensity, there is a need to implement layers that apply mathematical operations only at sparse tensor locations. It is also a prerequisite that these layers are differentiable in an end-to-end fashion.  

We use the Tensorflow library \cite{tensorflow2015-whitepaper} to implement the various sub-modules of our pipeline. We frequently utilize the \textit{scatter\_nd} operation, based on advantages mentioned in \cite{ren2018sbnet}, since it allows for the sparse update of pixel intensities at various tensor locations. We use this operation in the Bilinear Sampling Layer, such that, when interpolating pixel locations we only consider sparse neighbor locations of where depth intensities are available. 

We also contribute to an operation that prevents duplicate updates at the same pixel locations in eq. 5. In order to prevent rewriting pixel value updates to the same index locations, we use a Cantor Pairing function, and retain only unique index locations before updating depth intensity value.  

\subsection{Iterative Re-alignment}

So far we have presented a solution where an input mis-calibrated depth map is only transformed once before checking for photometric and distance errors. While some earlier works employ this method externally by resupplying the inputs ~\cite{schneider2017regnet}, our method is capable of applying iterative realignment in an end-to-end fashion within the network. For this, once the network predicts an initial transformation $T$, we transform the input mis-calibrated depth map by the predicted transformation. We now feed this newly transformed depth map as input to the network, in order to predict a residual transform $T'$. The final transformation is now computed after computing the product of predicted transformations at each stage. 

\begin{equation}
    \widehat{T} = (T)(T')(T'')(T''')...
\end{equation} 

At each step, the error is computed against the target depth map and target point cloud. When unrolled, the network resembles a recurrent unit similar to the work in \cite{newell2016stacked}, where the gradient flow at each iteration is against the transformed point cloud, and its resultant depth map.

%% file: 04results.tex
\section{EXPERIMENTS AND DISCUSSION}

We thoroughly analyze the proposed approach and present qualitative and quantitative results on LiDAR-camera data made available as part of the popular KITTI \cite{geiger2013vision} autonomous driving benchmark. In this section, we detail the experimental setup, training procedure, and analyze the results obtained.

\subsection{Dataset Preparation}

We use raw recordings from the KITTI dataset \cite{geiger2013vision}, specifically the color image and velodyne point cloud recordings, to prepare our dataset. We use the $26\_09$ driving sequences for training, since they consist of a high number of sequences with good scene variation. To obtain training data in the form of mis-calibrated depth maps, we first apply random transformations $T_{random}$ to the calibrated point clouds in order to decalibrate them. To capture a wide range of input misalignments, we sample $(v_x,v_y,v_z,\omega_x,\omega_y,\omega_z)$ randomly from a uniform distribution in the range of $\pm 10\degree$ rotation and $\pm 0.2$ m translation in any of the axes. We believe this is a good range of values to correctly emulate possible gross calibration errors that may not be directly corrected by target-based methods or measurement devices. By projecting these point clouds onto the camera plane, we obtain our input mis-calibrated sparse depth maps. The network takes as input these mis-calibrated depth maps and a corresponding RGB image.

\begin{figure}[!htb]
\begin{center}
    \setlength{\fboxsep}{2pt}%
    \setlength{\fboxrule}{1pt}%
    
    \includegraphics[width=1.0\linewidth]{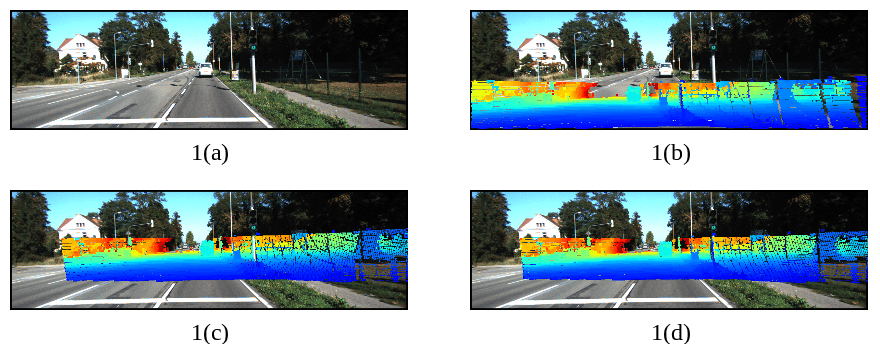}
\end{center}
    \caption{An example of a severely mis-calibrated input. Notice how the input mis-calibration causes the depth map to be skewed along the road. (a) is the RGB input image, and (b) is the mis-calibrated depth map. (c) is the calibrated depth map using our network's estimate, and (d) is the corresponding ground truth result. As shown, our network is able to recover the calibration even in this case.}
\label{fig:test}
\end{figure}

The target depth maps are obtained by recalibrating the misaligned depth maps. We apply the inverse of the original transform used to mis-calibrate the depth map, $T_{random}^{-1}$, and use this newly obtained depth map as the target depth map for photometric error reduction. While we could have directly used the calibrated point clouds to obtain ground truth depth maps during training, we noticed that for very large mis-calibrations, the point cloud often deviates outside the field-of-view of the camera, leading to huge variations between a calibrated depth map and a re-calibrated depth map. By this method, therefore, we ensure any there are no major differences between the predicted and ground truth maps due to wider transformations when calculating the photometric error.  

Using the above method, we generate a total of $24000$ pairs for training and $6000$ pairs for testing. To demonstrate the generalization capability of our architecture
even with different camera intrinsic parameters, we also use different instances from the $30\_09$ driving sequences during testing, as shown in Figure \ref{fig:test}.

\begin{figure*}[!htb]
\begin{center}
    \setlength{\fboxsep}{1pt}%
    \setlength{\fboxrule}{1pt}%
	\includegraphics[width=1.0\linewidth]{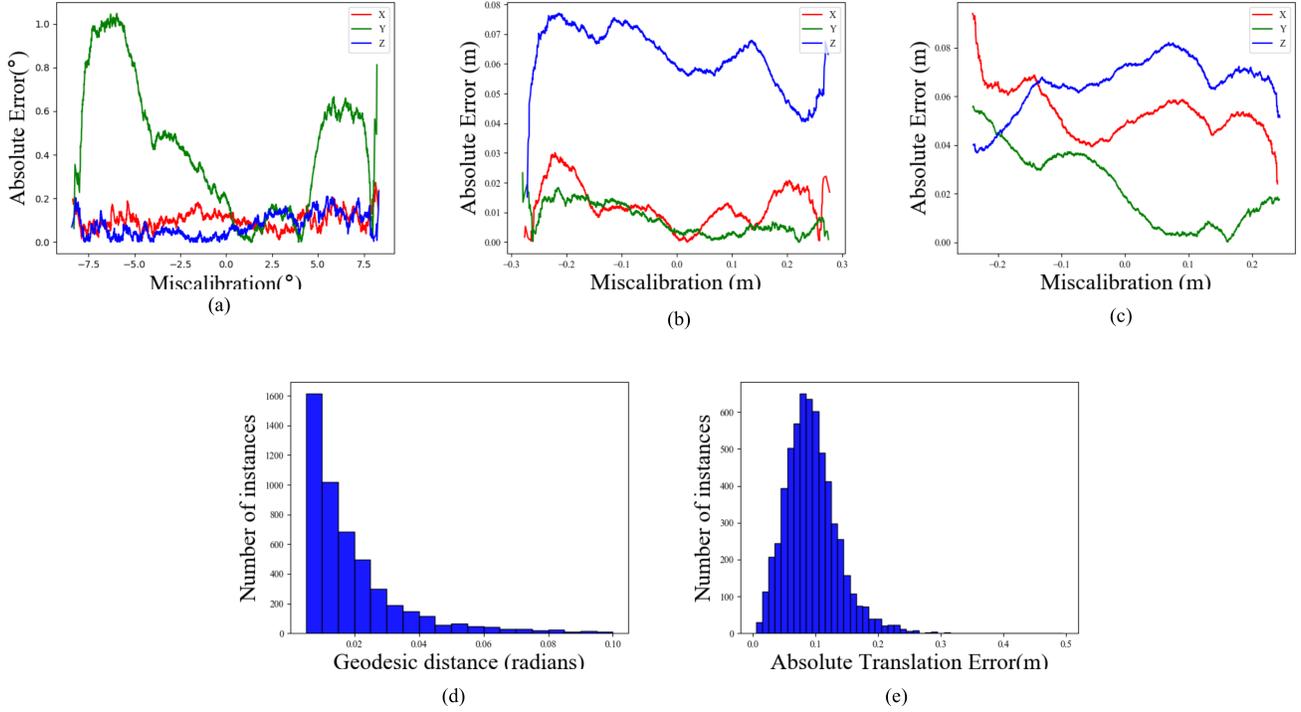}
    \caption{Global error statistics of our test set. Plots in red, green, and blue denote the absolute error values of \textit{CalibNet} along the X, Y, and Z axis respectively. (a) shows the change in absolute error for the range of initial mis-calibration in degrees across our test set. (b),(c) show the change in absolute error for the range of mis-calibration in meters. (b) demonstrates the error values upon training with ground truth rotations, while (c) demonstrates the error values when translation model is trained on prediction from the frozen weights of the previous iteration, (d) geodesic distance error distribution (in radians) over the test set instance pairs, (e) absolute translation error distribution (in meters) over the test set instance pairs.}
\label{fig:all_plots}
\end{center}
\end{figure*}

\subsection{Evaluation Metrics}


Note that, for the task of multi-sensor extrinsic calibration, while the translation measure between the camera and LiDAR coordinate frames can be roughly estimated using measuring devices, it is very difficult to even roughly measure the estimates of the yaw, pitch, and roll angles. Stressing on the difficulty of estimating the rotation components, we show a strong validation in using dense photometric error between the predicted and target depth maps to find the rotation matrix through our experiments. 

Since we are not training in a supervised manner, i.e. regressing by checking the error against the target transformation, we observed that while photometric loss helps in estimating better rotation values progressively over training, there is no such bound on the predicted translation values. This leads to erroneous initial translation values during training. We observed that point cloud distance measures serve as an important bound during training to ensure that translation estimates also slowly achieve the requisite target values. 
While we found the original end-to-end architecture effective in estimating the correct rotation values during testing, we found it difficult to estimate translation values, since residual values are hardly affected by small translation changes, but highly sensitive to erroneous rotation values. To remedy this, we train in an iterative fashion to re-estimate translation values. We experiment with freezing the initial model, and use the rotation predictions to transform the mis-calibrated point cloud. We then use only the point cloud distance measures to train for the translation values. We train this model with both ground truth rotations, and our estimated rotation values with frozen weights. We use earth mover's distance as our cloud distance metric of choice, since we observed that it scales better when estimating large translation values. Since the training time would increase drastically when calculating earth mover's distance for semi-dense point sets, we use the sparse depth maps without max-pooling to project sparse point clouds. We also further sparsify the clouds by finding centroids of local clusters, similar to the process mentioned for $d_{icp}$. We find $4096$ intermediate centroid locations in the predicted and ground truth point clouds, and use these centroids to calculate the earth mover's distance. We found this to improve training time without compromising on translation accuracy loss during training.

We evaluate overall rotation error as the geodesic distance over $SO(3)$, given by,
\begin{equation}
    d_g(R_i, R_j) = \frac{1}{\sqrt[]{2}}\|log(R_{i}^{T}R_j)\|_F
\end{equation}\\
and, absolute translation error as the overall error for translations,
\begin{equation}
    d_tr = \|X_i - X_j\|
\end{equation}\\
where $R_i$ is the predicted rotation, $R_j$ is the ground truth rotation, $X_i$ is the predicted translation, and $X_j$ is the ground truth translation.

\subsection{Training Details}

For training the network we use the Adam Optimizer \cite{DBLP:journals/corr/KingmaB14}, with an initial learning rate of $1e^{-4}$, and momentum equal to $0.9$. We decrease the learning rate by a factor $0.5$ every few epochs. We train for a total of $28$ epochs. Using Earth Mover Distance in the cost function, we set $\alpha_{ph}$ equal to $1.0$ and $\beta_{dist}$ equal to $0.15$ and slowly increase its value to $1.75$. To prevent over-fitting, we also apply a dropout of $0.7$ at the fully connected layers. 

    

\subsection{Results}

We show results of our base architecture for rotation values and the iterative re-alignment based model for translation.

\subsubsection{Rotation estimation}

Our network performs exceedingly well for rotation estimation. We report a mean absolute error (MAE) value for rotation angles on the test set: ({Yaw:} $0.15\degree$, {Pitch:} $0.9\degree$, {Roll:} $0.18\degree$). Figure \ref{fig:all_plots}(a) further illustrates the low absolute error values, against a widespread variation in mis-calibrations. We also show, in Figure \ref{fig:all_plots}(d), the performance of our network in predicting the overall rotation value as a function of the geodesic distance. Specifically, we demonstrate how the geodesic distance is close to $0$ for the bulk of the instance pairs in the test set.

\subsubsection{Translations given ground truth}

We observed that a single iteration fails to correctly estimate translation, since a significant photometric error reduction can lead to correctly estimated rotation parameters, but the point cloud distance loss has to decrease to very low values to correctly estimate translation. Since the value could not be minimized in a single iteration, we decided to further fine tune on translation values. During training for translation values, we rotate the input depth map with the ground truth rotation values and use the spatial transformer to apply the new estimate for translation. For this, we only use EMD as the error metric. We report mean absolute error in translation: ({X:} $4.2$ cm, {Y:} $1.6$ cm, {Z:} $7.22$ cm). Figure \ref{fig:all_plots}(b), and \ref{fig:all_plots}(d) correspond to the outputs based on this training scheme. Figure \ref{fig:all_plots}(b) shows the absolute error in translation over the mis-calibration range in the test set, while \ref{fig:all_plots}(d) shows the overall absolute translation error against a varying set, which, despite high initial mis-calibrations, is bounded in the range of cms. 

\subsubsection{Translations given CalibNet rotation estimates}

To demonstrate the capability of our geometric supervision method, we use the iterative re-alignment methodology in our network. After CalibNet regresses an initial estimate of rotation, we use the spatial transformer to apply the estimated rotation on the mis-calibrated depth map and train for translation using earth movers distance as the loss function. Figure \ref{fig:all_plots}(c) shows the overall absolute error for the various translation components. We observed that in the case of when mis-calibration in X-axis is higher than $\pm 0.25$ m, the absolute translation errors often deviate sharply. However, our model still works well for translation in the range of $\pm 0.24$ m. We report mean absolute error in translations: ({X:} $12.1$ cm, {Y:} $3.49$ cm, {Z:} $7.87$ cm).

\subsubsection{Qualitative Results}

Our network is able to accurately estimate the calibration, over a wide variety of scenes and wide range of initial mis-calibrations. Figures \ref{fig:long} and \ref{fig:test} show some of these results. Even with very high initial errors in all the $6$ axes, the spatial transformer successfully aligns the mis-calibrated depth map to the RGB frame, achieving close to ground truth re-alignment. Each of the columns corresponds to a particular scene chosen carefully to showcase the versatility of the network. For example the first column shows a scenario devoid of vehicles and less on features. The second column shows a scene with several vehicles and with changes in illumination in the scene. The third column shows a dense scene with a higher number of objects and low visibility of the underlying plane. In each of the cases, the network portrays its efficacy. More qualitative results are shown in the accompanying video.



%% file: 05conclusion.tex
\section{CONCLUSION}
In this paper, we presented a novel self-supervised deep network that can be used to estimate the $6$-DoF rigid body transformation between a 3D LiDAR and a 2D camera, a critical sensor combination that's typically found in most autonomous cars for perception. Unlike existing techniques, we do not rely on special targets in the scene, or any human intervention, thereby enabling true \textit{in-situ} calibration. The network is able to correct for mis-calibrations up to $\pm 20^{\circ}$ in rotation and $\pm 0.2$ m in translation, with a mean absolute error of $0.41^{\circ}$ in rotation and $4.34$ cm in translation.

The estimates of the network can serve as a good initialization for other optimization techniques with locally convex cost functions, which can further improve the estimates. Furthermore, the network seems to have learned the underlying geometry of the problem, which sets it up for other interesting applications in RGB-D localization, visual SLAM, stereo calibration, and is also a viable step in self-supervising architectures extended to non-symmetric inputs. In the future, we also wish to explore other priors to solve the registration problem between point clouds and RGB frames, such as the correspondence between depth maps and color frames, and ground-plane constraints. 